# A Literature Review: Stemming Algorithms for Indian Languages

M.Thangarasu[#1], Dr.R.Manavalan[*2]

[#] *Department of Computer Science and Application*
*K.S.Rangasamy College of Arts and Science*
*Tiruchengode, India*

*Abstract*— **Stemming is the process of extracting root word from the given inflection word. It also plays significant role in numerous application of Natural Language Processing (NLP). The stemming problem has addressed in many contexts and by researchers in many disciplines. This expository paper presents survey of some of the latest developments on stemming algorithms in data mining and also presents with some of the solutions for various Indian language stemming algorithms along with the results.**

## INTRODUCTION

A literature review is a description of the study relevant to particular field or topic. It is discusses about the information in a particular subject area for the past years. The process of reading, analyzing, evaluating and summarizing materials about specific topic are called Data Mining (DM), Text Mining (TM), Stemming and Clustering. The review describes summaries, evaluate and clarify various applications of clustering in different disciplines. This demonstrates that the knowledge has been gained in the required area and got awareness on the relevant theories.

### I. STEMMING: A REVIEW

Alkula, R. proposed plain character strings to meaningful words [1] in 2001. It is capable of improving the Information Retrieval System (IRS) using conversion of the plain text character strings to meaningful words. Krovetz.R viewed morphology as an inference process [2] in 1993. This research projected an inference process used in IRS. In 2002 Nilsson introduced hierarchical clustering using non-greedy principal direction divisive partitioning for partitioning algorithm is applied to number of clusters and it is based on the non-greedy principal. Popovic M and Willet.P presented the effectiveness of stemming for natural-language access to Slovene textual data [4] in 1992. It is also used to improve the stemming accuracy in Slovenes texts. In 1999 Savoy.J developed a stemming procedure and stop word list for general French corpora [5] for stems the French corpora using stop word list generation rule.

In the year of 1995 Kalamboukis.T.Z developed suffix stripping with Modern Greek [6]. In this research Greek language words are stemmed by suffix stripping algorithm. In 1999 Abu-Salem, H., Al-Omari, M., and Evens, M.W. applied Stemming methodologies over individual query words for an Arabic information retrieval system [7]. In the year of 2003 Rosell.M developed improving clustering of Swedish newspaper articles using stemming and compound splitting [8]. Morphological typology of languages for information retrieval [9] was invented by Pirkola.A. In this research, IRS retrieves the information based on the morphological typology. In the year 1996, Hull.D was developed stemming algorithms case study for detailed evaluation [10] for evaluating their performance.

### II. STEMMERS FOR INDIAN LANGUAGES: A REVIEW

In 2001, Shambhavi et al. introduced Kannada morphology analyzer and generator and using tire [11]. A lightweight stemmer for Hindi [12] was developed by Ramanathan et al. in the year of 2004. In this research, words conflate terms by suffix removal for information retrieval. Willet.P proposed the porter stemming algorithm for electronic library and information system [13] in 2006. Zahurul.MD et al. developed a lightweight stemmer for Bengali [14] in the year of 2009 for Bengali language spell checker. Assas-band, an affix-exception list based Urdu stemmer [15] was developed by Qurat-Ul-Ain Akram and et al. in the year of 2009. It stems the Urdu words using lexical lookup method (Assas-band). In 2010, Dinesh Kumar and Prince Rana developed design and development of stemmer for Punjabi [16], it uses Brute Force algorithm for stemming the Punjabi words.

Vijay Sundar et al. introduced Malayalam stemmer for information retrieval [17] in the year of 2010. Finite State Automata method is used to stem the Malayalam words.

TABLE I
STEMMER FOR INDIAN LANGUAGES

Tamil morphological analyzer [18] was introduced by Vijay Sundar et al. in the year of 2010. It is used to change the Tamil word to equal metrological Tamil words (lemmas).

| Method | Description | Author | Result | Year |
|---|---|---|---|---|
| Morphological Analyzer [11] | Derivational form of a word | Shambhavi et al. | 3700/88K | 2001 |
| Lightweight Algorithm [12] | To find the root word of a word | Ramanathaan et al. | 88% | 2004 |
| Porter Stemming Algorithm [13] | Remove inflectional endings form a word | Willett.P | 81.3% | 2006 |
| Lightweight algorithm [14] | To find the root word of a word | Zahurul Islam.Md et al. | 90.8% | 2009 |
| Assas-Band [15] | Stem the words based on exception list(lexical lookup) | Qurat-ul-Ain Akram et al. | 91.2% | 2009 |
| Brute Force Agorithm [16] | Truncate the derivational forms from a word | Dinesh Kumar et al. | 80.7% | 2010 |
| Finite State Automata (FSA) [17] | Transaction between the state(graph,table) | Vijay Sundar et al. | 95.8% | 2010 |
| Morphological Analyzer [18] | Derivational form of a word | Vijay Sundar et al. | 91.7% | 2010 |
| Suffix Stripping Algorithm [19] | Rules used for stems the words | Mudassar. M et al. | 82.5% | 2010 |
| Named Entity Recognition (NER) [20] | Identify the derivational form of a word | Sasidhar.B et al. | 94.1% | 2011 |
| Lightweight Algorithm [21] | To find the root word of a word | Juhi Ameta et al. | 91.5% | 2012 |
| MAULIK Algorithm [22] | Combination of Brute Force and Suffix Removal Algorithms | Upendra Mishra et al. | 91.8% | 2012 |
| Iterative Stemmer Algorithm [23] | Derivational form of word to a single root word | Vivek Anandan et al. | 84.3% | 2012 |

In 2010, discovering suffix a study for Marathi language [19] was proposed by Mudassar et al. for discovering hidden Marathi word in Knowledge Discovery Databases (KDD). Named entity in Telugu language using language dependent feature and rule based approach [20] was developed by Sridhar.B et al. in the year of 2011. The proposed model uses the Named Entity Recognition (NER) for stemming the Telugu words. Juhi Ameta et al. introduced a lightweight stemmer for Gujarati [21] in the year of 2012. In this proposed model, the lightweight stemmer algorithm for stems the Gujarati words. MAULIK: An efficient stemmer for Hindi [22] Language was developed by Upendra Mishra et al. MAULIK algorithm is used to stem the Hindi words. In 2012 an iterative stemmer for Tamil Language was proposed by Vivekanandan Ramachandran et al. in this proposed model, suffix stripper algorithm is used to stem Tamil words to its root word. The finding of the above literature survey is tabulated in Table I.

III. CONCLUSION

Stemming plays a vital role in information retrieval system and its effect is very large, compared to that found in review on various stemming algorithms. In this paper we have studied about various stemming algorithms and its effectiveness on various Indian languages. This is not quite enough for information retrieval system. So that in future, researchers will go for more number of implementations for the stemming algorithm methods and their utilities for various Indian languages information retrieval system.

## AUTHOR'S PROFILE

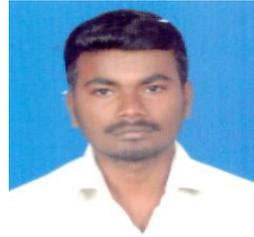

M.Thangarasu received Master of Computer Application degree from Bharathiar University. He purses M.Phil under supervision of Dr.R.Manavalan. His area of interest is data mining and Natural Processing Language (NLP).

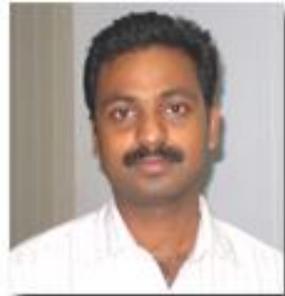

Dr. R. Manavalan is working as Assistant professor and Head in Department of computer science and Applications. He obtained Ph.D in Computer Science from Periyar University and published numerous research papers in international journals and also presented papers in various national and international conferences. His area of interest is soft computing, image processing and analysis, Theory of computation.